\pdfoutput=1

\PassOptionsToPackage{table}{xcolor}
\documentclass[11pt]{article}

\usepackage[final]{acl}

\usepackage{times}
\usepackage{latexsym}

\usepackage[T1]{fontenc}

\usepackage[utf8]{inputenc}

\usepackage{microtype}

\usepackage{inconsolata}

\usepackage{graphicx}
\usepackage{url}
\usepackage{tcolorbox}

\usepackage{mathtools}
\usepackage{amsmath, amssymb}
\usepackage{bbm}
\usepackage{booktabs}
\usepackage{tabularx}
\usepackage{multirow}
\usepackage{nccmath}  
\usepackage[table]{xcolor}
\usepackage{pgfplots}

\usepackage{pifont}
\newcommand{\cmark}{{\color{green!60!black}\ding{51}}} 
\newcommand{\xmark}{{\color{red}\ding{55}}}            

\title{Enhancing Study-Level Inference from Clinical Trial Papers via Reinforcement Learning-Based Numeric Reasoning}

\author{
 \textbf{Massimiliano Pronesti\textsuperscript{1,2}},
 \textbf{Michela Lorandi\textsuperscript{2}},
 \textbf{Paul Flanagan\textsuperscript{2}},
 \textbf{Oisín Redmond\textsuperscript{2}},
 \\
 \textbf{Anya Belz\textsuperscript{2}},
 \textbf{Yufang Hou\textsuperscript{1,3}}
\\
 \textsuperscript{1}IBM Research Europe - Ireland,
 \textsuperscript{2}Dublin City University,\\
 \textsuperscript{3}IT:U Interdisciplinary Transformation University Austria
\\
 \small{
   \textbf{Correspondence:} \href{mailto:massimiliano.pronesti@ibm.com}{massimiliano.pronesti@ibm.com}, \href{mailto:yufang.hou@it-u.at}{yufang.hou@it-u.at}
 }
}

\begin{document}
\maketitle


\begin{abstract}
Systematic reviews in medicine play a critical role in evidence-based decision-making by aggregating findings from multiple studies. A central bottleneck in automating this process is extracting numeric evidence and determining study-level conclusions for specific outcomes and comparisons. Prior work has framed this 
problem as a textual inference task by  
retrieving 
relevant content fragments and inferring conclusions from them. However, such approaches often rely on shallow textual cues and fail to capture the underlying numeric reasoning behind expert assessments.
In this work, we conceptualise the problem as one of quantitative reasoning. Rather than inferring conclusions from surface text, we extract structured numerical evidence 
(e.g., \emph{event counts} or \emph{standard deviations}) 
and apply domain knowledge informed logic to derive outcome-specific conclusions. We develop 
a numeric reasoning system composed of a numeric data extraction model and an effect estimate component,
enabling more accurate and interpretable 
inference aligned with the domain expert principles. 
%
We train the numeric data extraction model using different strategies, including supervised fine-tuning (SFT), 
and reinforcement learning (RL) with a new value reward model.
When evaluated on the \textsc{CochraneForest} benchmark, our best-performing approach --  using RL to train 
a small-scale  
number extraction model -- yields
up to a 21\% absolute improvement in F1 score over retrieval-based systems and outperforms general-purpose LLMs of over 400B parameters by up to 9\% on the RCTs benchmark.  Our results demonstrate the promise of reasoning-driven approaches for automating systematic evidence synthesis. 

\end{abstract}

\section{Introduction}
\label{sec:intro}

\begin{figure}[t!]
\centering
\includegraphics[width=.43\textwidth]{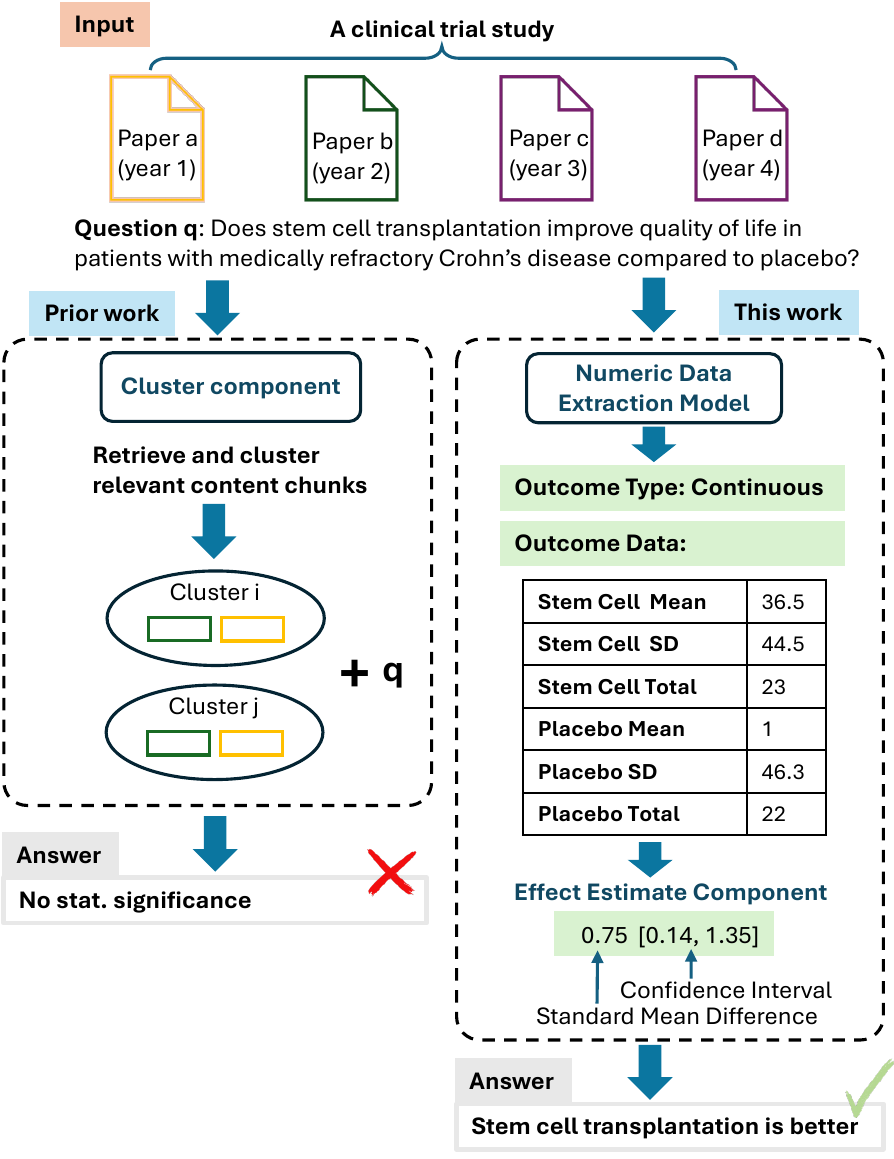}
\caption{Example of estimating the intervention effect based on the extracted outcome data for a clinical study.}
\label{fig:example}
\end{figure}

Systematic reviews are the cornerstone of evidence-based medicine, offering rigorous syntheses of available studies to guide clinical decision-making~\cite{murad2016new}. A critical component of systematic reviews is the extraction of study-level numeric evidence (e.g., \emph{event counts} or \emph{standard deviations}) from one or multiple corresponding clinical trial papers and derive the  conclusions for each outcome and comparison under assessment. However, automating this extraction remains an open challenge. As illustrated in Figure \ref{fig:example} (prior work), previous studies~\cite{pronesti2025}  have primarily framed this task as a retrieval-based question answering problem: given a query about an outcome, systems retrieve relevant study fragments and infer conclusions based on the retrieved text. 

Despite progress, such approaches fundamentally rely on surface-level textual cues, limiting their effectiveness.
To illustrate these limitations, we  plot in Figure~\ref{fig:ret_pre_cf} the relationship between evidence retrieval precision (x-axis) and the answer F1 score (y-axis) on 30 manually annotated instances (Appendix \ref{app:annot}) from the \textsc{CochraneForest} benchmark~\cite{pronesti2025} across four state-of-the-art large language models (LLMs). Each instance consists of a trial study comprising one or more papers, a research question, its corresponding categorical answer, and an exhaustive annotation of all supporting textual evidence drawn from the entire paper. In general, while better retrieval precision correlates with improved performance, even perfect retrieval (100\% precision) results in a modest maximum F1 score of 68\%. Moreover, performance gains plateau quickly: increasing retrieval precision from 50\% to 100\% yields only 3-4\% absolute improvement. This suggests that textual information alone is often insufficient to determine study conclusions, particularly when studies address multiple outcomes or consist of multiple publications.


\begin{figure} 
\centering
\resizebox{.4\textwidth}{!}{\begin{tikzpicture}
\begin{axis}[
    width=12cm,
    height=7cm,
    grid=both,
    xlabel={Retrieval Precision (\%)},
    ylabel={F1 Score},
    xmin=0, xmax=100,
    ymin=51.5, ymax=68.5,
    xtick={0,25,50,75,100},
    ytick={52,54,56,58,60,62,64,66,68},
    legend style={font=\small, legend columns=4, at={(0.47,1.04)}, anchor=south},
    line width=1pt,
    tick label style={font=\small},
    title style={font=\small},
    cycle list name=color list,
    legend style={draw=none}
]

\addplot+[mark=x, mark options={scale=1.2}, color=red] coordinates {
    (0,54) (25,60) (50,62) (75,62) (100,64)
};
\addlegendentry{GPT-4 (0125)}

\addplot+[mark=square*, mark options={scale=1.2}, color=cyan] coordinates {
    (0,52) (25,58) (50,60.5) (75,60) (100,63)
};
\addlegendentry{GPT-35-turbo (0125)}

\addplot+[mark=diamond*, mark options={scale=1.2}, color=orange!50] coordinates {
    (0,55) (25,59) (50,63) (75,63) (100,67.8)
};
\addlegendentry{LLaMa 3.3 70B}

\addplot+[mark=*, mark options={scale=1.2}, color=violet!40] coordinates {
    (0,58) (25,61) (50,61) (75,66) (100,68)
};
\addlegendentry{Qwen 2.5 72B}

\end{axis}
\end{tikzpicture}} 
\caption{F1 score for predicting the correct answers on 30 instances from \textsc{CochraneForest} on different retrieval precision using 4 state-of-the-art LLMs.} \label{fig:ret_pre_cf} 
\end{figure}


These findings motivate a shift in perspective -- from relying on surface-level textual cues to explicitly modelling the quantitative reasoning that underpins expert assessments in systematic reviews. A natural alternative is to adopt domain expert principles by extracting and interpreting the numerical evidence (e.g., \emph{effect sizes} and \emph{confidence intervals}) that supports each study’s conclusion. This reframes the task from semantic retrieval to structured statistical inference.
Recent work by~\citealp[]{yun2024automatically} has explored this direction 
by leveraging pretrained 
LLMs
through prompting to extract quantitative results. While promising, this approach has so far been limited to individual trials and has not addressed the challenges posed by longer, heterogeneous full-text studies included in systematic reviews. Furthermore, the potential of custom-trained models optimised specifically for numerical reasoning and alignment with expert conclusions remains largely unexplored in this setting.

Concurrently, advances in 
supervised fine-tuning (SFT) and 
reinforcement learning (RL) have shown significant improvements in aligning model behaviour with complex reasoning objectives~\cite{weifinetuned, guo2025deepseek}. Building on these insights, as shown in Figure \ref{fig:example} (this work), 
we propose a structured pipeline that extracts interpretable numerical evidence from full-text studies via a numeric data extraction model, and infers study-level conclusions using transparent rules through an effect estimate component—eschewing reliance on implicit textual signals. We train  compact numeric data extraction models using a wide range of strategies, including SFT, SFT with intermediate reasoning traces, and RL
with a novel value-based reward model. On two different datasets, our models outperform the prompting-based approaches based on big models proposed in \newcite{yun2024automatically}.

In addition, compared to the implicit text evidence reasoning approach \cite{pronesti2025},  our models allow one to automatically generate the corresponding row of a forest plot (Section \ref{sec:forest_plot}) from a full text study by directly extracting numerical evidence for each outcome measure. This represents a key step toward full automation of the systematic review process, bridging the gap between primary study reporting and meta-analytic synthesis \citep{wallace2010semi, tsafnat2014systematic}.

In summary, our contributions are as follows:
(1) we propose a novel pipeline that predicts study conclusions by extracting and statistically analysing numerical outcomes, instead of relying purely on textual retrieval; (2) we explore how different training strategies—standard SFT, SFT with intermediate reasoning traces, and RL—affect the reasoning abilities of compact language models on this task; (3) we develop custom models trained for this task, achieving up to 21\% absolute improvement in F1 score compared to retrieval-based systems on \textsc{CochraneForest}.

\section{Preliminaries}

\subsection{Systematic Reviews}
A systematic review is a rigorous method of synthesising evidence from multiple studies that address a clearly formulated research question~\cite{cochranehandbook}. It follows a structured protocol for identifying, selecting, and appraising relevant research to minimise bias and yield reliable findings.

\subsection{Forest Plots}
\label{sec:forest_plot}
Forest plots are visual tools commonly used in systematic reviews to display the estimated effects from multiple studies on a common scale. Each study is typically represented as a point estimate (e.g., \emph{mean difference}, \emph{odds ratio}) with a confidence interval, and a vertical line indicating a null effect (e.g., 0 for differences, 1 for ratios). Forest plots help in assessing consistency across studies and interpreting the overall effect direction and magnitude. Examples are shown in Appendix~\ref{sec:forest_plots_appendix}.

\subsection{Data Extraction from Biomedical Studies}
Biomedical studies, particularly randomised controlled trials (RCTs), typically report numerical outcome data for different treatments or interventions. These results—such as event or group size—may appear in tables, figures, or embedded within narrative text. Extracting this information is essential for downstream tasks such as study-level evidence synthesis and across-study meta-analysis.

Following \newcite{yun2024automatically}, we categorise extracted numerical data into two main outcome types: \emph{binary outcomes}, which include the number of events and group sizes for both the intervention and comparator arms; and \emph{continuous outcomes}, which consist of means, standard deviations, and group sizes for intervention and comparator groups.


Once extracted, these values can be used in standard meta-analytic methods to estimate treatment effects—such as mean differences, risk ratios, or odds ratios—along with their corresponding confidence intervals, forming the basis for deriving study-level conclusions. An example of estimating treatment effects is provided in Appendix~\ref{sec:forest_plots_appendix}.

\section{Methodology}
As shown in Figure \ref{fig:example}, our system consists of a fine-tuned numeric extraction model grounded in reasoning, alongside a rule-based effect size estimation component. In the following, we describe each part in detail.

\subsection{Training Dataset Creation}\label{dataset}
\paragraph{Training Data Collection.}
The initial phase of our methodology involved the acquisition of high-quality, human-annotated data to train our models. Following the methodology described in~\citet{pronesti2025}, we processed the Cochrane Database of Systematic Reviews (CDSR)\footnote{\url{https://www.cochranelibrary.com/cdsr/reviews}} to identify systematic reviews containing non-paywalled full-text studies and at least one forest plot available in SVG format. Each forest plot was parsed to extract the underlying numerical data (i.e. the number of events and total participants in each group for binary outcomes, or the mean, standard deviation, and sample size for continuous outcomes), along with the point estimate, 95\% confidence interval, and textual conclusion for each included study. 

To increase dataset size and training diversity, we relaxed one of the constraints imposed in the original \textsc{CochraneForest} benchmark. Specifically, we no longer require that each forest plot contain at least two studies with differing conclusions. This change allows us to include more reviews while still preserving outcome-level heterogeneity across the dataset. In addition, to prevent data leakage, we explicitly excluded any study in \textsc{CochraneForest} from our training set. 

The final training dataset consists of 2,072 examples, spanning 104 systematic reviews and 25.9 M tokens of full-text biomedical content. Each data point contains: (1) the full text of a study, (2) the outcome type (i.e., \emph{binary} or \emph{continuous}) as defined in the corresponding forest plot, (3) the set of numerical values to be extracted (e.g., group means or event counts), (4) the computed point estimate, (5) the 95\% CI, and (6) the final conclusion assigned to the outcome in the forest plot. We refer to this dataset as \textsc{CochraneForestExt}.

\begin{table}[t]
\scalebox{0.75}{
\centering
\begin{tabular}{lcccccc}
\toprule
\textbf{Dataset} & \textbf{Train} & \textbf{Test} & \textbf{Total}  & \textbf{Avg tokens}\\
\midrule
\textsc{CochraneForestExt} & 1864 & 208 & 2072  & 12109.2 \\
\textsc{CochraneForest} & – & 725 & 725 & 11688.7 \\
\textsc{RCTs} & – & 413 & 413  & 4364.9 \\
\bottomrule
\end{tabular}
}
\caption{Datasets statistics. Train/test split only applies to \textsc{CochraneForestExt}. \textsc{CochraneForest} and RCTs are used for testing.}
\label{tab:dataset_stats}
\end{table}


\paragraph{Synthetic Data Annotation.}
To 
enrich the dataset with reasoning traces for SFT, we used Llama 3.1 405B~\cite{grattafiori2024llama} with the system prompt shown in Figure~\ref{fig:sys_prompt} (Appendix), temperature of 0.7 and 2,048 tokens generation limit. An example data instance is provided in Table~\ref{tab:data_example} (Appendix).

\subsection{Numeric Data Extraction Model}
\subsubsection{SFT with CoT}

We first adapt a pretrained language model to our task via supervised fine-tuning (SFT), which adapts a pretrained language model $\pi_\theta$ to reflect a domain-specific distribution $\mathcal{P}$. This is achieved by minimising the negative log-likelihood over a dataset of example sequences, encouraging the model to increase the probability of desired outputs. In our context, the goal is to enable the model to map free-text descriptions of study results to structured outputs that capture outcome information in a standardised schema. Each training example consists of a biomedical study, paired with the corresponding outcome of interest (Table \ref{tab:data_example}). The output is represented in YAML format and encodes either binary outcomes (with event counts and totals for intervention and comparator groups) or continuous outcomes (with group-wise means, standard deviations, and sample sizes). The model learns to generate these structured summaries conditioned on the corresponding textual evidence. That is, given prompt-target pairs $(\mathbf{x}, \mathbf{y})\sim \mathcal{P}$, the loss is computed as:
\[
\mathcal{L}_{\text{cond}}(\theta) = - \mathbb{E}_{(\mathbf{x}, \mathbf{y}) \sim \mathcal{P}} \left[ \sum_{t=1}^{m} \log \pi_\theta(y_t \mid \mathbf{x}, y_{<t}) \right]
\]

This focus helps the model better learn task-relevant outputs without overfitting to input tokens~\cite{wang-etal-2023-self-instruct,vicuna2023, yu-etal-2024-lions}.

\subsubsection{RL with Fine-grained Rewards}


As an alternative to SFT, we explore Reinforcement Learning, which further refines LLMs by aligning model outputs with human preferences or reward signals. We adopt Group-Relative Policy Optimisation (GRPO)~\cite{grpo}, which computes normalised rewards over a group of responses and reduces variance in learning.

In our setup, the model acts as a policy $\pi_\theta$ that takes as input a textual passage describing clinical study results and outputs a structured response. Each response consists of a thought process enclosed in a \texttt{<think>} tag, followed by a YAML object encoding outcome data, using the same schema adopted in SFT. For each passage $\mathbf{x}$, $G$ candidate completions $\{y_i\}_{i=1}^G \sim \pi_\text{old}( \cdot \mid \mathbf{x})$ are sampled from the reference policy $\pi_\text{old}$ to encourage robustness and diversity. These completions are scored using rule-based reward functions that evaluate factual correctness and adherence to format. The raw rewards $R_i$ are then normalised across the group:
\[
A_i = \frac{R_i - \mathbb{E}[R_j]}{\sqrt{\mathbb{V}[R_j]}}, \quad j \in \{i,...,G\}
\]
where $\mathbb{E}[R_j]$ and $\mathbb{V}[R_j]$ are respectively the mean and variance of the rewards for the group of responses. The policy is optimised using a clipped, KL-regularised objective that encourages agreement with high-reward behaviours while maintaining proximity to a reference model $\pi_{\mathrm{ref}}$:

{\footnotesize
	\begin{align*}
	\mathcal{L}_{\text{GRPO}}(\theta) =\;&
		\mathbb{E} \bigg[ \frac{1}{G} \sum_{i=1}^G \frac{1}{|y_i|} \sum_{t=1}^{|y_i|} 
		\min
        \Big( p_{i,t}(\theta) A_i,\,\\
		&\quad \hspace{-1cm}
		\text{clip}(p_{i,t}(\theta), 1-\varepsilon,1+\varepsilon) A_i \Big) \notag  - \beta\, \text{KL}[\pi_\theta \,\|\, \pi_{\text{ref}}] \bigg]
	\end{align*}
}


where $\beta$ governs the regularisation strength and  $p_{i,t}(\theta)$ is the token-level probability ratio defined as follows: 
\[
p_{i,t}(\theta) = \frac{\pi_\theta (y_{i,t} \mid x, y_{i,<t})}{\pi_{\theta_\text{old}} (y_{i,t} \mid x, y_{i,<t})}
\]

\paragraph{Reward Functions.}

In our setting, the model produces structured YAML outputs representing binary or continuous outcomes. To evaluate these outputs during RL, we define three rule-based reward functions based on format validity and numerical correctness. Let $C_i$ be the model output, $E_i$ the expected answer.

\paragraph{Correctness Reward (CR).}  
This reward compares numerical values in $C_i$ and $E_i$ when the outcome type is correct. Let $V_i = \{v_1, \dots, v_n\}$ and $\hat{V}_i = \{\hat{v}_1, \dots, \hat{v}_n\}$ be the parsed numerical fields. Then:
\[
R_{\mathrm{CR}} = \frac{1 + \sum_{j=1}^n \mathbbm{1}\{v_j \approx \hat{v}_j\}}{1 + n}
\]
where $v_j \approx \hat{v}_j$ means exact match for integers and absolute difference $< 10^{-3}$ for floats. If parsing fails or types mismatch, we set $R_{\mathrm{CR}} = 0$.

\paragraph{Format Reward (FR).}  
This reward checks whether $C_i$ follows the expected structure. Let $\mathcal{F}$ denote the set of all valid formats (Appendix~\ref{app:prompts}), including required keys and outcome type. We define:
\[
R_{\mathrm{FR}} = 
\begin{cases}
1 & \text{if } \pi_{\theta}(x) \in \mathcal{F} \\
0 & \text{otherwise}
\end{cases}
\]
This reward ensures that the output adheres to a valid YAML schema for the predicted outcome type.

\paragraph{Thought Format Reward (TFR).}
The TFR incentivises the model to adhere to a predefined output structure, such as the use of \texttt{<think>} tag.
\[
R_{\mathrm{TFR}} = 
\begin{cases}
1 & \text{if } \pi_{\theta}(x) \text{ matches thought pattern}  \\
0 & \text{otherwise}
\end{cases}
\]

\paragraph{Final Reward.}
The final reward is a weighted combination of format and correctness components:
\[
R = 0.8 \cdot R_{\mathrm{CR}} + 0.1 \cdot R_{\mathrm{FR}} + 0.1 \cdot R_{\mathrm{TFR}}
\]
The weights prioritise factual accuracy while still incentivising structural correctness and robustness to formatting issues with proper reasoning traces. 
More details are provided in Appendix~\ref{app:hyper_and_apis}.

\subsection{Effect Estimate Component}

After extracting the relevant numerical data, we compute standardised fixed-effect estimates based on the type of outcome~\cite{hedges1998fixed}.

\paragraph{Binary Outcomes.} 
For event-based outcomes, we compute the risk ratio (RR) as \(\text{RR} = \frac{a/(a+b)}{c/(c+d)}\), where \(a, b\) are the numbers of events and non-events in the treatment group, and \(c, d\) in the control group. To quantify uncertainty, we compute the 95\% confidence interval on the log scale as \(\log(\text{RR}) \pm 1.96 \cdot \sqrt{\frac{1}{a} - \frac{1}{a+b} + \frac{1}{c} - \frac{1}{c+d}}\), which is then exponentiated to return to the RR scale (see Appendix \ref{sec:forest_plots_appendix}  for an example calculation).

\paragraph{Continuous Outcomes.} 
For outcomes measured on a continuous scale, we compute the mean difference (MD) as \(\bar{x}_T - \bar{x}_C\), where \(\bar{x}_T\) and \(\bar{x}_C\) are the group means in the treatment and control arms, respectively. The 95\% CI is then computed as $\text{MD} \pm 1.96 \cdot \sqrt{\frac{s_T^2}{n_T} + \frac{s_C^2}{n_C}}$, where \(s_T, s_C\) and \(n_T, n_C\)  are respectively the standard deviations and samples sizes of the two groups (see Appendix \ref{sec:forest_plots_appendix}  for an example calculation).

\paragraph{Deriving Study Conclusions.}  
Study conclusions are determined directly from the 95\% confidence interval of the effect estimate~\cite{5minmeta}. For binary outcomes, if the confidence interval for the odds ratio lies entirely above 1, the study is classified as supporting the intervention; if it lies entirely below 1, it favors the control; and if it includes 1, the result is considered inconclusive. For continuous outcomes, the same logic applies with respect to the null value 0.

\section{Experiments}
\subsection{Experimental setup}
\paragraph{Training and Evaluation Datasets.} For training, we use the dataset created as described in Section~\ref{dataset}, with the reasoning traces for SFT, using 1864 samples for training and 208 for validation.

For evaluation, we use two datasets. The first is \textsc{CochraneForest}~\cite{pronesti2025}, which consists of 725 instances derived from 48 Cochrane Systematic Reviews and 220 forest plots. The second, introduced by \citet{yun2024automatically}, contains 413 complete, human-annotated instances derived from 120 RCTs. Dataset statistics are reported in Table~\ref{tab:dataset_stats}.

\begin{table*}[t]
	\small
	\centering
	\renewcommand{\arraystretch}{1.2}
	\setlength{\tabcolsep}{4pt}
	\begin{tabular}{lc|cccccc|cccccc}
		\toprule
		\multirow{2}{*}{\textbf{Model}} & \multirow{2}{*}{\textbf{Think}} & \multicolumn{6}{c|}{\cellcolor{gray!20}\textsc{\textbf{CochraneForest}}} & \multicolumn{6}{c}{\cellcolor{gray!20}\textbf{RCTs}~\cite{yun2024automatically}} \\
		& & Acc & F1 & EM & EM@1 & MSE$\downarrow$  & EIR$\downarrow$ & Acc & F1 & EM & EM@1 & MSE$\downarrow$  & EIR$\downarrow$\\
		\toprule
		\multicolumn{2}{c}{\cellcolor{gray!20}\textbf{Pretrained LLMs}} & \multicolumn{6}{c}{\cellcolor{gray!20}} & \multicolumn{6}{c}{\cellcolor{gray!20}} \\
		\texttt{GPT-4-0125} & \xmark & 71.4 & 73.7 & 28.4 & 72.1 & 0.88 & 0.31  & 71.2 & 65.3 & 34.5 & 70.3 & 0.69 & 0.44\\
		\texttt{Qwen2.5-7B} & \xmark & 69.7 & 65.4 & 14.9 & 66.4 & 1.40  & 0.39 &  63.2 & 59.7 & 20.3 & 73.8 &  1.19 & 0.50 \\
		\texttt{Qwen2.5-14B} & \xmark &  76.2& 73.7 &  27.4 & 73.6 &0.72  & 0.30 & 69.8 & 66.4 & 28.6 & 75.8 &  0.91 & 0.44 \\
		\texttt{Qwen2.5-72B} & \xmark & 82.2 & 80.3 & 42.5 & 81.7 & 0.63 & 0.26 &  \textbf{75.3}&  \textbf{72.8}& \textbf{35.1} &\textbf{81.1} & \textbf{0.68} & \textbf{0.39}\\
		\texttt{Llama-3.1-8B} & \xmark & 68.7 & 66.6 & 15.9 & 62.2 & 2.16 & 0.33 & 62.7 & 54.7 & 14.8 & 66.5 & 1.30 & 0.47 \\
		\texttt{Llama-3.1-70B} & \xmark & 79.4 & 78.3 & 31.5 & 78.7 & 0.59 & 0.28 & 65.5 & 60.8 & 30.0 & 74.2 & 1.45 & 0.44\\
		\texttt{Llama-3.1-405B} & \xmark & \textbf{82.4} &\textbf{80.8} & \textbf{44.3} & \textbf{82.4} & \textbf{0.43}  & \textbf{0.23} & 70.3 & 66.5 & 33.6 & 76.7 & 0.80 & 0.41 \\
		\texttt{DeepSeek-Qwen-7B}& \cmark & 59.7 & 54.3 & 11.2 & 58.3& 2.83 &  0.50& 53.8 & 45.7  & 9.4 &  53.8 &  3.68 & 0.61  \\
		\texttt{DeepSeek-Qwen-14B} & \cmark & 65.4  & 61.2 & 19.9 &  66.4 & 1.29 & 0.44 &60.2  & 55.1  & 11.2 & 60.3 & 2.41 & 0.58 \\
		\texttt{DeepSeek-Qwen-32B} & \cmark & 74.0 & 71.6 & 28.6 & 73.5 &  0.58 & 0.28 & 68.8 & 65.0 & 29.1 & 78.2 & 0.72 &0.43\\
		\toprule
		\multicolumn{2}{c}{\cellcolor{gray!20}\textbf{Our Models}} & \multicolumn{6}{c}{\cellcolor{gray!20}} & \multicolumn{6}{c}{\cellcolor{gray!20}} \\
		\texttt{Qwen2.5-7B-SFT} & \cmark & 74.5  & 70.1 & 28.4 & 79.1 & 0.51 &  0.29&  71.5& 68.4 & 30.2 & 80.0&  0.64 & 0.36  \\
		\texttt{Qwen2.5-7B-RL} & \cmark & \textbf{81.6} & \textbf{80.1} & \textbf{42.2} & \textbf{81.4}  & \textbf{0.40} & \textbf{0.24}  & \textbf{79.3} & \textbf{76.4} & \textbf{41.2} & \textbf{89.7} & \textbf{0.49} &\textbf{0.28}\\
		\bottomrule
	\end{tabular}
	\caption{Evaluation results across models on two datasets. We report Accuracy and F1 (label prediction), EM, EM@1, EIR and MSE (numerical extraction).}
	\label{tab:results}
\end{table*}

\paragraph{Evaluation Metrics.} 
We report a range of metrics to assess both end-to-end performance and intermediate reasoning accuracy. For the end-to-end task of predicting the final conclusion label of each study, we report accuracy and F1 score. Following \citet{yun2024automatically}, we also evaluate the quality of numerical extraction using exact match (EM), EM@1 (at least one positional match), as well as the mean squared error (MSE) of the computed point estimate derived from the extracted values. In addition, we measure the Error Impact Rate (EIR), defined as the ratio between 
the number of extraction errors that lead to a flipped conclusion and the number of total extraction errors at the study level.
\[
\text{EIR} = \frac{\# \text{errors that flip the conclusion}}{\# \text{extraction errors}}
\]
This metric provides an insight into how often extraction mistakes materially affect downstream decision-making. A high EIR indicates that even a few extraction errors can substantially alter the final prediction, whereas a low EIR suggests that the model's conclusions are more robust to imperfect inputs. Thus, EIR serves as a proxy for understanding the correlation between intermediate numerical accuracy and end-to-end reliability.

\paragraph{Training Setup.} 
We conduct our training using the Qwen2.5 model family~\cite{qwen2.5}, specifically the 7B variant. Two distinct training regimes are explored: SFT and RL. In the results section, models are labelled with subscripts corresponding to the respective training strategy.

SFT is performed for 5 epochs with a batch size of 1 using a learning rate of $5 \times 10^{-5}$ and the AdamW optimiser~\cite{adamw}. For the RL setup, we adopt the GRPO algorithm~\cite{grpo}, training for 3 epochs with a learning rate of $1 \times 10^{-6}$, batch size 1, and 16 sampled generations per batch. 
Additional details are provided in Appendix~\ref{app:hyper_and_apis}.

\paragraph{Model Baselines.}
To validate our results, we compare a range of open- and closed-source models,
with and without reasoning capabilities. All models are evaluated in zero-shot settings with prompt and hyperparameters shown in Appendix~\ref{app:prompts}
. 

We include two main model families: Qwen 2.5~\cite{qwen2.5} and Llama 3.1~\cite{grattafiori2024llama}. In addition, we benchmark DeepSeek-R1 and the distilled Qwen models derived from it. We exclude distilled models afferent to the Llama family because of their limited context size. 
For closed-source models, we use \texttt{GPT-4-0125}.

In addition, we compare performances on the end-to-end task against the two best RAG baselines for this task: URCA~\cite{pronesti2025}, which clusters retrieved passages based on their embedding vectors and filters relevant information from each cluster given the query; and GraphRAG~\cite{graphrag}, which builds a graph-based text index by summarising closely related entities from the source documents.


\subsection{Main Results} 
\paragraph{Comparison with Pretrained Baselines.}
Table~\ref{tab:results} presents a performance comparison of pretrained and fine-tuned language models on the \textsc{CochraneForest} and RCTs datasets. Models vary in size, architecture, and training approach, allowing us to examine the relationship between model scale, reasoning capability, and task-specific performance. Among the pretrained baselines, \texttt{Llama-3.1-405B} achieves the best performance on \textsc{CochraneForest},  closely followed by \texttt{Qwen2.5-72B}, which performs competitively on both datasets, topping the RCTs benchmark on several metrics. \texttt{Llama-3.1-70B} and \texttt{Qwen2.5-14B} also rank highly, while \texttt{Qwen2.5-7B} and \texttt{Llama-3.1-8B} underperform relative to their larger variants. 

Fine-tuning
improves performance significantly. \texttt{Qwen2.5-7B-SFT} outperforms all pretrained models of similar or larger size. The \texttt{Qwen2.5-7B-RL} model establishes a new state-of-the-art, surpassing all baselines—including the 405B model—in nearly every metric on RCTs and closely matching it on \textsc{CochraneForest}, despite being nearly 58x smaller. In addition, when compared directly to its pretrained counterpart, \texttt{Qwen2.5-7B-RL} shows large gains in EM (+20.9), accuracy (+16.1), and F1 (+16.7) on the RCTs dataset, highlighting the effectiveness of RL.

Notably, we observe that reasoning capabilities do not always lead to improved performance. In fact, all the distilled DeepSeek models with reasoning perform worse than their non-reasoning counterparts. These results suggest that general reasoning ability alone is insufficient for complex domain-specific tasks. Instead, explicit task supervision and structured reasoning training appear necessary to guide models in applying reasoning capabilities effectively.

\paragraph{Comparison with RAG Baselines.}  
Table~\ref{tab:rag_number_baselines} compares the best-performing numbers-based models with two strong retrieval-augmented generation (RAG) baselines ---URCA and GraphRAG---on the \textsc{CochraneForest} dataset.  Both RAG methods underperform compared to direct numerical reasoning. The fine-tuned \texttt{Qwen2.5-7B-RL} achieves an absolute gain of over 21 F1 points compared to the best RAG baseline, highlighting the limitations of retrieval in settings where precise numerical grounding and structured inference are required. Even the zero-shot numbers-based approach outperforms the RAG methods, suggesting that factual retrieval alone is insufficient without robust reasoning over structured content.

\begin{table}[ht]
\small
	\centering
	\begin{tabular}{l|cc}
		\toprule
		\textbf{Method} & \textbf{Acc} & \textbf{F1} \\
		\midrule
		\rowcolor{gray!15}
		\multicolumn{3}{l}{\textbf{RAG baselines}} \\
		URCA~\cite{pronesti2025}   &    60.2 & 58.8        \\
		GraphRAG~\cite{graphrag}   &   59.6 & 57.5        \\
		\midrule
		\rowcolor{gray!15}
		\multicolumn{3}{l}{\textbf{Numerical baselines}} \\
		Qwen2.5-7B (zero-shot)  &    69.7    &  65.4      \\
		Qwen2.5-7B-SFT        &  74.5      & 70.1       \\
		Qwen2.5-7B-RL         &    \textbf{81.6}    &  \textbf{80.1}      \\
		\bottomrule
	\end{tabular}
	\caption{Performance comparison between RAG-based and numbers-based approaches on \textsc{CochraneForest} using \texttt{Qwen2.5-7B-Instruct}.}
	\label{tab:rag_number_baselines}
\end{table}

\subsection{Ablation Studies}

To assess the impact of different input modalities and training strategies, we conduct a series of ablation experiments (Table~\ref{tab:ablations}) on the RL- and SFT-trained models. These include both data ablations and training ablations.

In the data ablations, we isolate the contribution of different components of the input. We evaluate model performance when provided with: (i) only textual context, where all tables are removed; (ii) only tables, where we exclude surrounding text and provide the model with structured numerical content; and (iii) only the top retrieved chunks obtained with URCA~\cite{pronesti2025}, restricting access to 10 evidence passages per query.


In the training ablations, we analyse the effect of intermediate reasoning and reward design. To evaluate the role of reasoning traces, we compare the SFT model to a version trained without Chain-of-Thought (CoT), which predicts the final answer directly without intermediate steps. To assess the impact of the reward function, we compare the RL model to another version trained using the \texttt{EX} reward, which only provides a positive signal when all predicted values are correct. 

\[
\scalebox{1}{$
R_{\mathrm{EX}} = \mathbbm{1}\left\{ \bigwedge_{j=1}^n \left( v_j \approx \hat{v}_j \right) \right\}
$}
\]

We observe that among the data ablations, the models trained only on tables achieve the best performance, confirming the importance of structured numerical evidence in this task. However, there is a consistent drop in performance across all settings when models only have access to a single input type compared to the full input (text + tables), indicating that both textual and tabular information are necessary for accurate reasoning. The \texttt{URCA} setting, which corresponds to the output of a retrieval-based pipeline, yields the lowest scores. This aligns with the findings in Table~\ref{tab:rag_number_baselines} and further highlights the limitations of existing retrieval-augmented generation approaches in this setting, primarily due to their inability to recover the precise numerical content required for grounded inference.

In the training ablations, removing CoT supervision during SFT (\texttt{SFT-no-CoT}) results in lower performance compared to full SFT, showing that intermediate reasoning traces help guide the model’s learning process. When using RL, we find that the dense reward signal ($R_{\mathrm{CR}}$) significantly outperforms the variant trained with the EX reward ($R_{\mathrm{EX}}$). This confirms that sparse rewards are less effective at shaping model behaviour than denser, fine-grained signals. 

		

\begin{table}[t]
\small
	\centering
	\begin{tabular}{ll|cc}
		\toprule
		\textbf{Model} & \textbf{Input} & \textbf{Acc} & \textbf{F1} \\
		\midrule
		\rowcolor[gray]{0.9} \multicolumn{4}{l}{\textbf{Input Data Ablations}} \\
		\multirow{3}{*}{\texttt{Qwen2.5-7B-SFT}} & \texttt{text} & 62.3 & 60.1 \\
		 & \texttt{tables}  & 70.4 & 66.5 \\
		 & \texttt{urca}  & 56.7 & 54.0 \\
		\bottomrule
		\multirow{3}{*}{\texttt{Qwen2.5-7B-RL}} & \texttt{text}  & 65.8 & 63.2 \\
		 & \texttt{tables}  &\textbf{ 73.1} & \textbf{71.6 }\\
		 & \texttt{urca}  & 59.2 & 56.8 \\
		\bottomrule
		
		\rowcolor[gray]{0.9} \multicolumn{4}{l}
		{\textbf{Training Ablations}} \\
		\texttt{Qwen2.5-7B} & \texttt{text + tables} & 69.7    &  65.4  \\
		\texttt{+ SFT-no-CoT} & \texttt{text + tables} &71.2 & 68.3 \\
        \texttt{+ SFT}
        \textbf{(ours)} & \texttt{text + tables} &  74.5      & 70.1  \\
        \texttt{+ RL-EX} & \texttt{text + tables} & 79.7 & 78.8\\
		\texttt{+ RL} \textbf{(ours)} & \texttt{text + tables} &    \textbf{81.6}    &  \textbf{80.1} \\
		
		\bottomrule
	\end{tabular}
	\caption{Ablation study on model inputs and training supervision on \textsc{CochraneForest}. \texttt{urca} refers to the top 10 chunks retrieved by its namesake RAG approach.}
	\label{tab:ablations}
\end{table}

\subsection{Analyses}
\begin{table*}[h]
	 \small
	 \centering
	\begin{tabular}{
			c >{\raggedright\arraybackslash}p{5.8cm}
			>{\centering\arraybackslash}m{1.2cm} >{\centering\arraybackslash}m{1.5cm}
			>{\centering\arraybackslash}m{1.2cm} >{\centering\arraybackslash}m{1.5cm}
		}
		\toprule
		\textbf{Thought Process} & \textbf{Reasoning Label}
		& \multicolumn{2}{c}{\texttt{Qwen2.5-7B-RL}} 
		& \multicolumn{2}{c}{\texttt{Qwen2.5-7B-SFT}} \\
		\cmidrule(lr){3-4} \cmidrule(lr){5-6}
		& & Count & Fraction & Count & Fraction \\
		\midrule
		\textbf{\cmark} & Correct reasoning with traceability          & 42 & 31.82\%  & 34 & 25.76\% \\
		\textbf{\cmark} & Correct reasoning without traceability       & 39 & 29.55\% & 24 & 18.18\%  \\
		\textbf{\cmark} & Correct reasoning, incorrect number          & 37 & 28.03\% & 41  & 31.06\% \\
		\textbf{\xmark} & Correct number, incorrect reasoning          & 3 & 2.72\% & 13 & 9.85\%  \\
		\textbf{\xmark} & Copy without reasoning                       &  6&4.55\%  & 0 &  0.00\%\\
		\textbf{\xmark} & Hallucinated                                 &5  &3.79\%  & 20  & 15.15\% \\
		\textbf{\xmark} & Missing reasoning                            & 0 & 0.00\% & 0 & 0.00\% \\
		\midrule
		& \textbf{Total}                                               & 132 & 100\% & 132 & 100\% \\
		\bottomrule
	\end{tabular}
	\caption{Distribution of annotated reasoning labels for \texttt{Qwen2.5-7B-RL} and and {\texttt{Qwen2.5-7B-SFT}. \textbf{\cmark} indicates correct reasoning. \textbf{\xmark} indicates incorrect reasoning.}}
	\label{tab:hum_eval_labels_models}
\end{table*}

\paragraph{Impact of the Thought Process.} We want to evaluate whether the reasoning processes generated by our trained models (RL and SFT) are logically sound, provide meaningful explanations for the extracted numbers, and enable traceability back to the input papers. To this end, we manually annotate the outputs of both models on 30 biomedical studies, for which human annotators had previously identified the sources of the correct numbers (see Appendix \ref{app:annot}), 
assigning a single label to each reasoning process. Labels distinguish whether the reasoning correctly supports the individual number, whether it enables traceability to the source, and whether it hallucinates or lacks explanation. A complete list of labels and annotation criteria is provided in Appendix~\ref{app:annot_thought}.

Results (Table~\ref{tab:hum_eval_labels_models}) show that the RL model produces a higher fraction of correct reasoning processes overall compared to the SFT model (61.37\% vs 43.94\%, summing the first two rows). This suggests that reinforcement learning improves the model’s ability to provide plausible and well-structured explanations for numerical outputs. Notably, the RL model achieves a substantially lower rate of hallucinated reasoning (3.79\% vs 15.15\%), highlighting a significant improvement in factual grounding. It also shows a reduced frequency of incorrect reasoning behind correct numbers (2.72\% vs 9.85\%), indicating better alignment between the generated explanations and the numerical outputs.


These findings suggest that reinforcement learning not only improves the factual accuracy of model outputs, but also enhances the traceability and reliability of the underlying reasoning process.

\paragraph{Reward Dynamics.} 
Figure~\ref{fig:training_curves} shows the reward dynamics during RL training. Thought Format Reward (TFR) and Format Reward (FR) increase rapidly and plateau early, indicating that the model quickly learns the reasoning scaffold and YAML schema. This behaviour also suggests that the model inherently learns to predict the correct outcome type at an early stage. By contrast, Correctness Reward (CR) improves more gradually and continues rising after TFR/FR have stabilised, showing that numerical correctness requires longer training. These dynamics are consistent with our main results and design choices: structural aspects are mastered quickly, whereas correctness is more demanding and depends on the model’s ability to apply reasoning across each component of the final output. Overall, the reward dynamics illustrate how reinforcement signals guide the model from surface-level alignment toward substantive reasoning ability.

\begin{figure}
    \centering
    \resizebox{\columnwidth}{!}{
\begin{tikzpicture}
  \begin{axis}[
    width=12cm,
    height=8cm,
    grid=both,
    xlabel={Step},
    ylabel={Training Reward},
    xmin=0, xmax=705,
    ymin=0.1, ymax=1.02,
    legend style={font=\small, legend columns=3, at={(0.49,1.02)}, anchor=south}
  ]
  \addplot[cyan, mark=*, thick, smooth] coordinates {
(1,0.8594)
(5,0.7930)
(10,0.7562)
(15,0.8094)
(20,0.8812)
(25,0.8281)
(30,0.8719)
(35,0.9313)
(40,0.9594)
(45,1.0000)
(50,0.9812)
(55,0.9812)
(60,0.9594)
(65,0.9844)
(70,0.9781)
(75,0.9656)
(80,0.9812)
(85,0.9875)
(90,0.9906)
(95,1.0000)
(105,0.9953)
(110,0.9938)
(115,0.9156)
(120,0.9812)
(125,0.9969)
(130,0.9969)
(135,0.9969)
(140,1.0000)
(145,0.9906)
(150,0.9938)
(155,0.9906)
(160,0.9750)
(165,0.9906)
(170,0.9906)
(175,0.9875)
(180,0.9563)
(185,0.9969)
(190,0.9875)
(195,0.9844)
(205,0.9859)
(210,0.9938)
(215,0.9938)
(220,0.9938)
(225,0.9969)
(230,0.9938)
(235,1.0000)
(240,0.9969)
(245,1.0000)
(250,0.9969)
(255,0.9875)
(260,0.9719)
(265,1.0000)
(270,0.9969)
(275,0.9969)
(280,1.0000)
(285,1.0000)
(290,1.0000)
(295,0.9969)
(305,0.9969)
(310,1.0000)
(315,0.9969)
(320,1.0000)
(325,0.9938)
(330,0.9969)
(335,0.9875)
(340,0.9750)
(345,0.9875)
(350,0.9812)
(355,0.9812)
(360,0.9906)
(365,0.9969)
(370,0.9938)
(375,1.0000)
(380,0.9969)
(385,0.9969)
(390,1.0000)
(395,1.0000)
(405,1.0000)
(410,1.0000)
(415,0.9969)
(420,1.0000)
(425,1.0000)
(430,1.0000)
(435,0.9906)
(440,0.9969)
(445,0.9969)
(450,1.0000)
(455,0.9969)
(460,0.9969)
(465,1.0000)
(470,0.9938)
(475,0.9969)
(480,1.0000)
(485,1.0000)
(490,1.0000)
(495,0.9875)
(505,0.9984)
(510,0.9938)
(515,0.9938)
(520,1.0000)
(525,0.9938)
(530,1.0000)
(535,1.0000)
(540,1.0000)
(545,0.9969)
(550,1.0000)
(555,1.0000)
(560,0.9906)
(565,0.9969)
(570,0.9969)
(575,0.9844)
(580,0.9969)
(585,0.9969)
(590,0.9969)
(595,1.0000)
(605,1.0000)
(609,1.0000)
(615,1.000)
(620,1.000)
(625,1.000)
(630,1.000)
(635,1.000)
(640,1.000)
(645,1.000)
(650,1.000)
(655,1.000)
(660,1.000)
(665,1.000)
(670, 0.998)
(675, 0.996)
(680,1.000)
(685,1.000)
(690,1.000)
(695,1.000)
};\addlegendentry{YAML Format Reward}

\addplot[red, mark=triangle*, thick, smooth] coordinates {
(1,0.8644)
(5,0.9161)
(10,0.9606)
(15,0.9875)
(20,0.9906)
(25,0.9969)
(30,0.9906)
(35,1.0000)
(40,0.9938)
(45,1.0000)
(50,1.0000)
(55,1.0000)
(60,1.0000)
(65,0.9969)
(70,1.0000)
(75,0.9938)
(80,0.9938)
(85,0.9906)
(90,0.9969)
(95,1.0000)
(105,0.9984)
(110,1.0000)
(115,1.0000)
(120,0.9969)
(125,0.9969)
(130,1.0000)
(135,0.9969)
(140,0.9969)
(145,0.9969)
(150,1.0000)
(155,0.9969)
(160,1.0000)
(165,1.0000)
(170,0.9938)
(175,0.9969)
(180,0.9688)
(185,1.0000)
(190,1.0000)
(195,1.0000)
(205,1.0000)
(210,1.0000)
(215,1.0000)
(220,1.0000)
(225,1.0000)
(230,1.0000)
(235,0.9969)
(240,1.0000)
(245,1.0000)
(250,1.0000)
(255,1.0000)
(260,0.9969)
(265,1.0000)
(270,1.0000)
(275,1.0000)
(280,0.9969)
(285,0.9969)
(290,1.0000)
(295,0.9938)
(305,0.9984)
(310,0.9969)
(315,1.0000)
(320,1.0000)
(325,1.0000)
(330,1.0000)
(335,1.0000)
(340,1.0000)
(345,1.0000)
(350,1.0000)
(355,1.0000)
(360,0.9969)
(365,1.0000)
(370,0.9969)
(375,0.9969)
(380,1.0000)
(385,1.0000)
(390,1.0000)
(395,1.0000)
(405,1.0000)
(410,1.0000)
(415,1.0000)
(420,1.0000)
(425,1.0000)
(430,1.0000)
(435,0.9969)
(440,1.0000)
(445,1.0000)
(450,1.0000)
(455,1.0000)
(460,1.0000)
(465,1.0000)
(470,1.0000)
(475,1.0000)
(480,1.0000)
(485,1.0000)
(490,1.0000)
(495,1.0000)
(505,1.0000)
(510,1.0000)
(515,1.0000)
(520,1.0000)
(525,1.0000)
(530,1.0000)
(535,1.0000)
(540,1.0000)
(545,1.0000)
(550,1.0000)
(555,1.0000)
(560,1.0000)
(565,1.0000)
(570,1.0000)
(575,1.0000)
(580,1.0000)
(585,1.0000)
(590,1.0000)
(595,1.0000)
(605,0.9984)
(609,1.0000)
(615,1.000)
(620,1.000)
(625,1.000)
(630,1.000)
(635,1.000)
(640,1.000)
(645,1.000)
(650,1.000)
(655,1.000)
(660,0.9993)
(665,1.000)
(670,1.000)
(675,0.9976)
(680,1.000)
(685,1.000)
(690,1.000)
(695,1.000)
};\addlegendentry{Thought Format Reward}

\addplot[green!80, mark=o, thick, smooth] coordinates {
(1,0.1215)
(5,0.1724)
(10,0.2521)
(15,0.2998)
(20,0.2637)
(25,0.3549)
(30,0.3319)
(35,0.3028)
(40,0.3259)
(45,0.4000)
(50,0.3243)
(55,0.3651)
(60,0.4070)
(65,0.4005)
(70,0.3481)
(75,0.3951)
(80,0.4793)
(85,0.4419)
(90,0.4681)
(95,0.4886)
(100, 0.4685)
(105,0.4785)
(110,0.5323)
(115,0.5276)
(120,0.5000)
(125,0.5391)
(130,0.5262)
(135,0.4860)
(140,0.5769)
(145,0.4919)
(150,0.5917)
(155,0.4940)
(160,0.4583)
(165,0.5629)
(170,0.4831)
(175,0.5911)
(180,0.4958)
(185,0.5404)
(190,0.5511)
(195,0.6029)
(205,0.5632)
(210,0.5243)
(215,0.5570)
(220,0.5883)
(225,0.6109)
(230,0.6012)
(235,0.5945)
(240,0.5572)
(245,0.5904)
(250,0.5993)
(255,0.6207)
(260,0.6139)
(265,0.6685)
(270,0.6823)
(275,0.6074)
(280,0.6527)
(285,0.6565)
(290,0.6994)
(295,0.7024)
(305,0.6857)
(310,0.7064)
(315,0.6857)
(320,0.6441)
(325,0.6720)
(330,0.6612)
(335,0.6428)
(340,0.6343)
(345,0.6725)
(350,0.7039)
(355,0.6475)
(360,0.6672)
(365,0.6853)
(370,0.6596)
(375,0.7341)
(380,0.6655)
(385,0.7462)
(390,0.6946)
(395,0.6327)
(405,0.7342)
(410,0.6387)
(415,0.7199)
(420,0.7679)
(425,0.6242)
(430,0.7678)
(435,0.7149)
(440,0.7103)
(445,0.7044)
(450,0.6863)
(455,0.6883)
(460,0.6734)
(465,0.7209)
(470,0.7066)
(475,0.6965)
(480,0.7340)
(485,0.7106)
(490,0.6813)
(495,0.6901)
(505,0.7322)
(510,0.6831)
(515,0.6887)
(520,0.7473)
(525,0.6791)
(530,0.7447)
(535,0.7202)
(540,0.7171)
(545,0.7335)
(550,0.6262)
(555,0.7428)
(560,0.6954)
(565,0.7442)
(570,0.6659)
(575,0.7292)
(580,0.7905)
(585,0.7017)
(590,0.7792)
(595,0.7983)
(605,0.8224)
(609,0.7587)
(615, 0.7862)
(620, 0.7723)
(625, 0.7342)
(630, 0.7285)
(635, 0.7151)
(640, 0.7923)
(645, 0.7889)
(650, 0.7601)
(655, 0.7702)
(660, 0.7384)
(665, 0.7221)
(670, 0.7503)
(675, 0.7935)
(680, 0.7556)
(685, 0.7483)
(690, 0.7365)
(695, 0.7242)
};\addlegendentry{Correctness Reward}
  \end{axis}
\end{tikzpicture}}
    \caption{Rewards dynamics. Format rewards plateau quickly during training, whereas the correctness reward improves more gradually, suggesting that LLMs naturally learn to follow structural patterns while still improving on quality rewards.}
    \label{fig:training_curves}
\end{figure}

\paragraph{Qualitative Example.}
We present a qualitative example from \textsc{CochraneForest} (Figure~\ref{fig:qualitative_example}, in appendix), comparing the reasoning and outputs of \texttt{Qwen2.5-7B-SFT} and \texttt{Qwen2.5-7B-RL} on a challenging case where the exact values for the comparator and outcome are not explicitly reported in the biomedical study. While the SFT model fails to extract the relevant information, the RL model demonstrates stronger reasoning capabilities: it identifies population percentages from one of the study’s tables and correctly maps them to the total number of participants previously extracted, thereby analytically inferring the desired values.  Notably, even our human annotators initially struggled to locate the correct numbers in this scenario. 

\section{Related work}

\paragraph{Biomedical Information Extraction.}
Recent efforts in biomedical information extraction have made significant advances.  \citet{wadhwa2023jointlyextractinginterventionsoutcomes} introduced a generative approach for extracting interventions, comparing and outcomes from RCTs. While effective, we extend this by focusing on numerical evidence and reasoning. 

\citet{lehman-etal-2019-inferring} demonstrated that accurately extracting evidence is the primary bottleneck for determining treatment efficacy. \citet{odoherty-etal-2024-beyond} showed using abstracts alone rather than structured PICO data can worsen synthesis quality, underscoring the importance of data representation. Our model processes the full content of the paper, providing a more comprehensive basis. Our work can be seen as an example of scientific argument mining at the global discourse level \cite{al-khatib-etal-2021-argument} and contributes to the broader agenda of AI for Science \cite{eger2025transformingsciencelargelanguage}.

\paragraph{LLMs for Evidence Synthesis and Numerical Reasoning.}
While LLMs have shown promise in biomedical text mining, they face challenges synthesising complex evidence and numerical reasoning. \citet{shaib-etal-2023-summarizing} found that GPT-3 struggled with multi-document synthesis and biased effect reporting. \citet{nye-etal-2020-trialstreamer} introduced a framework to map evidence and infer conclusions. \citet{yun2024automatically} assessed how effectively LLMs can extract numerical data, noting challenges with continuous outcomes and distinguishing similar measures. \citet{wang2025accelerating} introduced a method that integrates LLMs with code generation to extract clinical study outcomes. However, their evaluation was restricted to selected cancer-related reviews, and the reliance on structured prompts and domain-specific heuristics raises questions about scalability to broader therapeutic areas. \citet{lai2025language} evaluated LLMs for data extraction and risk of bias assessment in complementary medicine, reporting high performance, but their focus on a relatively narrow and domain-specific set of trials limits the generalizability of the findings. Similarly, \citet{sun2024good} assessed LLMs for automated data extraction from randomized trials, finding encouraging results, yet performance was uneven across outcome types, with continuous measures proving particularly challenging. These studies highlight that while LLMs show high potential, existing work focuses on narrow domains, curated benchmarks, or simplified tasks, leaving open the question of how well they generalise to the diverse evidence encountered in real-world systematic reviews.


\paragraph{Reinforcement Learning for Numerical Evidence Extraction.}
Previous work has proven that SFT can be used to improve models such as BERT \cite{devlin2019bertpretrainingdeepbidirectional} for biomedical text mining \cite{Lee_2019, XIE2022109460}.
Reinforcement learning has improved model alignment for complex reasoning objectives \citep{weifinetuned, lambert2024tulu, guo2025deepseek}.  Recent work~\cite{lin2025ehrmind} demonstrates that RL with verifiable rewards can improve model performance on clinical reasoning tasks like EHR-based calculations and trial matching. To the best of our knowledge, we are the first to design and apply RL-based approaches specifically to the extraction of numerical evidence from biomedical studies for use in systematic reviews.


\section{Conclusion}
In this paper we presented a quantitative reasoning framework for automating evidence extraction in systematic reviews, shifting away from shallow textual inference toward structured numeric understanding. By directly modelling the process domain experts follow,
we move closer to automating a key step in evidence-based medicine. Our proposed system, combining SFT and RL numeric extraction models with reasoning over effect estimates, significantly outperforms retrieval-based baselines and even large-scale language models on both the \textsc{CochraneForest} and the RCTs benchmarks. These results affirm the value of reasoning-driven supervision for complex scientific tasks and point to RL as a promising avenue for developing reliable and domain-aligned extraction systems in evidence-based medicine. 

Beyond improving performance, our approach highlights the importance of aligning machine reasoning with expert workflows, ensuring interpretability and trustworthiness. Looking forward, this work opens opportunities for integrating automated evidence extraction into clinical decision support pipelines, ultimately reducing the manual burden on researchers and accelerating the translation of medical knowledge into practice. Future research may extend this framework to other scientific domains where quantitative reasoning is core. 

\section{Limitations}
While this work makes meaningful contributions toward reasoning-based evidence extraction, several limitations remain. First, the system assumes that all relevant numerical evidence is explicitly or implicitly reported and cleanly extractable, which is often not the case in real-world clinical studies. Missing values and inconsistent formats can hinder reliable extraction. At present, the model does not attempt to identify or flag missing information, which limits its utility in incomplete or noisy settings. Training models to detect and explicitly report missing or uncertain values is a critical direction for future development. Additionally, our approach currently focuses on a limited set of outcome types and uses predefined logical rules, which may constrain generalisation to more complex or diverse clinical scenarios. Addressing these limitations will be key to deploying robust systems for real-world systematic review automation.

\bibliography{custom}

\appendix

\section{Forest Plots and Effect Size Estimation}
\label{sec:forest_plots_appendix}

Forest plots are a standard way to visualise the results of individual studies and their synthesis in a meta-analysis. Each row typically represents one study, showing its estimated treatment effect and the corresponding 95\% confidence interval (CI). The point estimate is usually plotted as a square (with size proportional to the study’s weight), and the CI as a horizontal line. The vertical line represents the line of no effect: 1 for ratios (e.g., risk ratio or odds ratio), and 0 for differences (e.g., mean difference).

At the bottom of the plot, a diamond often represents the pooled effect estimate from all included studies. This visualisation helps readers assess not only the overall direction and magnitude of the effect but also the consistency (heterogeneity) across studies.

\begin{figure}[h]
\includegraphics[width=\columnwidth]{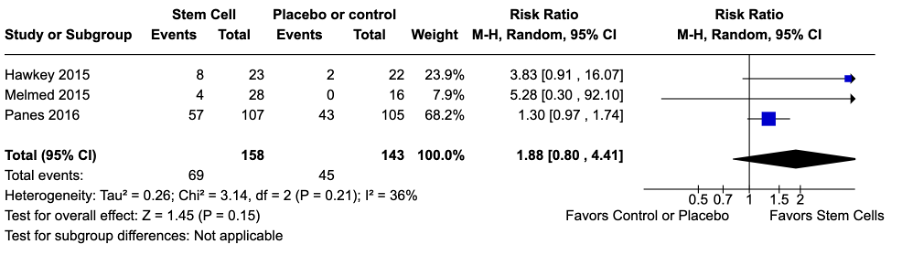}
\caption{A forest plot assessing clinical remission (binary outcome) in patients affected by medically refractory Crohn’s disease treated with stem cells versus control.}
\label{fig:fp_bin}
\end{figure}

\begin{figure}[h]
\includegraphics[width=\columnwidth]{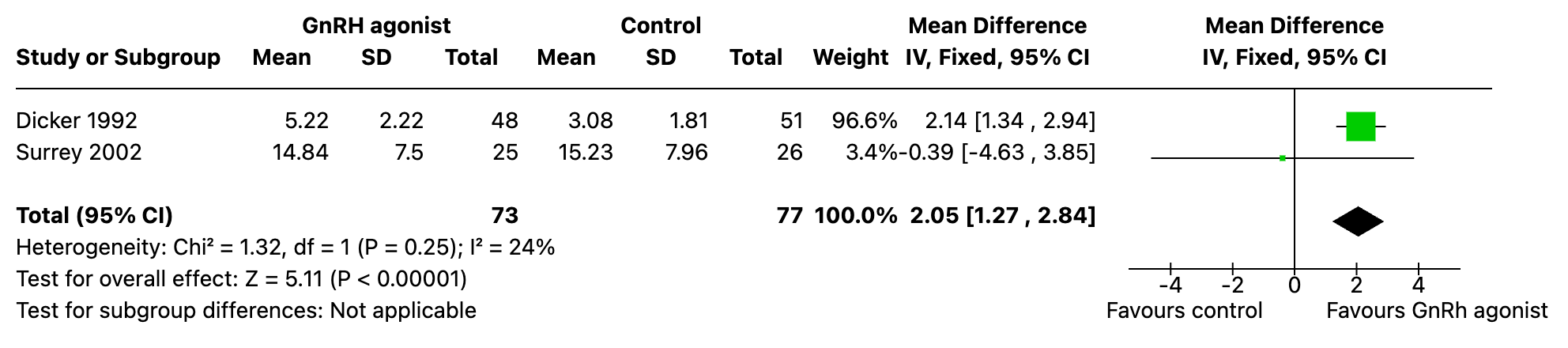}
\caption{A forest plot comparing GnRH agonist versus no agonist (placebo) in women with endometriosis, assessing a continuous outcome (average number of oocytes per woman).}
\label{fig:fp_cont}
\end{figure}

Below we illustrate how to interpret and compute effect estimates and confidence intervals for the two common outcome types: binary outcomes with risk ratios (Figure~\ref{fig:fp_bin}), and continuous outcomes with mean differences (Figure~\ref{fig:fp_cont}).

\paragraph{Binary Outcomes and Risk Ratios.}
Binary outcomes refer to variables with two possible states, such as event/no event. A commonly used effect measure in this context is the \emph{risk ratio} (RR), which compares the proportion of events in the intervention group to that in the comparator group:

\[
\text{RR} = \frac{a/n_1}{c/n_2}
\]

where $a$ and $n_1$ are the number of events and total participants in the intervention group, and $c$ and $n_2$ are those in the comparator group.

\paragraph{Example.} In the \textbf{Hawkey 2015} study (Figure~\ref{fig:fp_bin}):

\begin{itemize}
	\item Intervention (stem cells): 8 events out of 23 participants
	\item Comparator (placebo): 2 events out of 22 participants
\end{itemize}

The point estimate is:

\[
\text{RR} = \frac{8/23}{2/22} \approx \frac{0.3478}{0.0909} \approx 3.83
\]

To compute the 95\% confidence interval, we use the standard error of 
$\log$:~

\[
\text{SE} = \sqrt{\frac{1}{8} - \frac{1}{23} + \frac{1}{2} - \frac{1}{22}} \approx \sqrt{0.536} \approx 0.732
\]

The CI on the log scale is:

\[
\log(3.83) \pm 1.96 \cdot 0.732 \approx (-0.091, 2.777)
\]

Exponentiating the bounds gives the 95\% CI for the RR:

\[
\text{95\% CI} \approx (e^{-0.091}, e^{2.777}) \approx (0.91, 16.07)
\]

\paragraph{Continuous Outcomes and Mean Differences.}
Continuous outcomes are measured on a numerical scale, such as a score or a lab value. The typical effect measure is the \emph{mean difference} (MD), which is the arithmetic difference in average outcome values between groups:

\[
\text{MD} = \bar{x}_1 - \bar{x}_2
\]

where $\bar{x}_1$ and $\bar{x}_2$ are the group means.

\paragraph{Example.} In the \textbf{Dicker 1992} study (Figure~\ref{fig:fp_cont}):

\begin{itemize}
	\item Intervention (GnRH agonist): mean = 5.22, SD = 2.22, $n=48$
	\item Comparator (control): mean = 3.08, SD = 1.81, $n=51$
\end{itemize}

The point estimate is:

\[
\text{MD} = 5.22 - 3.08 = 2.14
\]

The standard error is computed as:

\[
\text{SE} = \sqrt{\frac{2.22^2}{48} + \frac{1.81^2}{51}}  \approx 0.408
\]

Then, the 95\% confidence interval is:

\[
2.14 \pm 1.96 \cdot 0.408 \approx 2.14 \pm 0.80 = (1.34, 2.94)
\]

\section{Prompts}\label{app:prompts} The prompts used for synthetic data annotation and for training are shown in Figure~\ref{fig:sys_prompt} and~\ref{fig:train_inf_prompt}, respectively. For training, a temperature of 0.7 and 2,048 tokens as maximum output length are used.

\section{Manual Annotations Identifying Number Sources}\label{app:annot}

\begin{figure*}[h!]
 \centering
\includegraphics[width=1\textwidth]{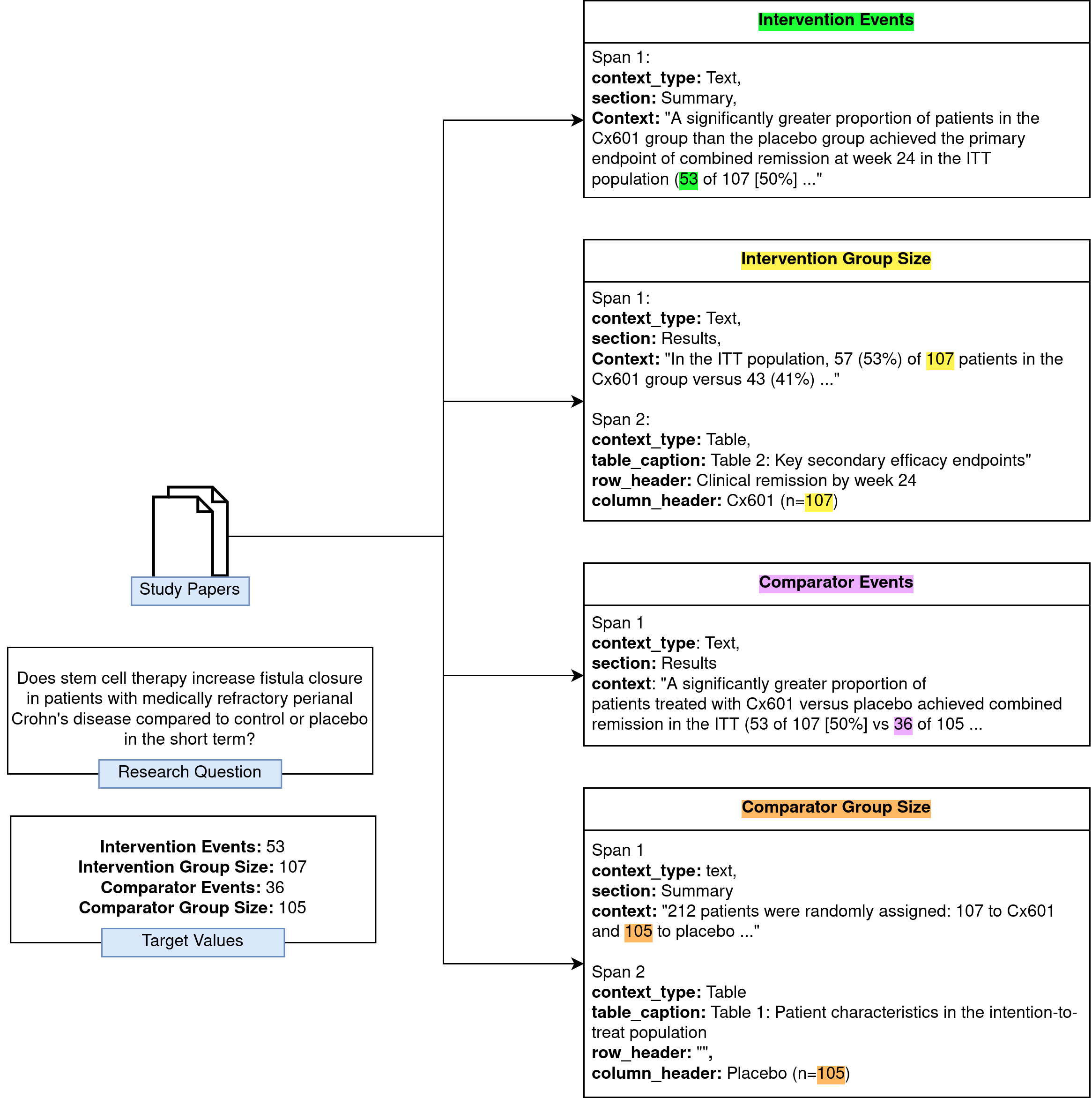}
\caption{Example of multi-span numerical data annotations for a given study}
\label{fig:annot_span}
\end{figure*}

\paragraph{Annotation Instances.}
Two master's degree students in NLP and co-authors of this work annotated 
30 
studies paired with outcomes from the \textsc{CochraneForest} dataset. These studies were randomly sampled from the whole dataset, and each student annotated 
fifteen studies with an overlap of five studies for cross-comparison. For each study, the annotators were instructed to locate and mark all textual or tabular spans that contained the numerical evidence supporting the reported outcomes. In the case of binary outcomes, this involved identifying four values: the event count and total group size for both the intervention and comparator arms. For continuous outcomes, six values were annotated: the mean, standard deviation, and group size for each group. Each annotated number was linked to one or more corresponding spans from the original study, enabling fine-grained traceability.
An example annotation instance is shown in Figure~\ref{fig:annot_span}.

\paragraph{Inter-annotator Agreement.}
To assess annotation consistency, we computed inter-annotator agreement (IAA) on the five studies annotated by both annotators. Agreement was evaluated at the span level: a match was counted when both annotators identified overlapping text spans referring to the same numerical value (e.g., intervention group size or mean outcome). We report a span-level F1 score of 0.70, indicating substantial agreement. Disagreements were primarily due to differences in span boundaries or the inclusion of surrounding contextual phrases.

\begin{figure}
    \small
    \begin{tcolorbox}[
            colback=gray!10, 
            colframe=black, 
            boxrule=0.5mm,
            title={Prompt for synthetic data annotation}, 
            fonttitle=\bfseries 
        ]
            \{study\_content\}
            \\
            \\
            The above is a study from a medical systematic review.
            Your task is to produce a reasoning to explain how to extract the relevant numbers for the following intervention, comparator and outcome:
            \\
            \\
            intervention: \{intervention\}\\
            comparator: \{comparator\}\\
            outcome: \{outcome\} \\
            \\
            The expected output is:
            \\
            \\
            \{target\_value\}
            \\
            \\
            You need to first explain how to infer the outcome type.
            Then, which numbers to extract and why. Finally, point to where these numbers appear in the study.
        \end{tcolorbox}
        \caption{Prompt for the synthetic data annotation with reasoning traces.}
        \label{fig:sys_prompt}
\end{figure}

\begin{figure}
	\small
	\begin{tcolorbox}[
		colback=gray!10, 
		colframe=black, 
		boxrule=0.5mm,
		title={Prompt for training and inference}, 
		fonttitle=\bfseries 
		]
		Articles: \{articles\}
		\\
            \\
		Question: Based on the given trial articles, what is the outcome type and corresponding numerical data for the following Comparison and Outcome?
		\\
		\\
		Comparison: \{comparison\}\\
		Outcome: \{outcome\}
		\\
		\\
		First, determine and output the outcome\_type as either: binary or continuous
		\\
		\\
		Then, provide the extracted data in format as follows:
		\\
		If the outcome is binary, use this format:
		\\
		\\
		outcome\_type: binary
		\\intervention:
		\\events: NUMBER total: NUMBER
		\\comparator:
		\\events: NUMBER total: NUMBER
		\\
		\\
		If the outcome is continuous, use this format:
		\\
		\\
		outcome\_type: continuous
		\\intervention:
		\\mean: NUMBER standard\_deviation: NUMBER group\_size: NUMBER
		\\
		comparator:
		\\mean: NUMBER standard\_deviation: NUMBER group\_size: NUMBER
		\\
		\\Use post-intervention data when both pre and post are available. If multiple timepoints are reported, choose the one closest to the timepoint of interest, or the latest available.
		Think about it step by step.
	\end{tcolorbox}
	\caption{Prompt for training and inference.}
	\label{fig:train_inf_prompt}
\end{figure}

\section{Manual Annotations  for Extraction Model Outputs}\label{app:annot_thought}
After some pilot studies, we use the following labels to categorize a model's thinking process:
\begin{enumerate}
    \item \textit{Correct reasoning with traceability}, the reasoning correctly supports the generated numbers and provides sufficient clues to identify their location in the input papers;
    \item \textit{Correct reasoning without traceability}, the reasoning correctly supports the generated numbers, but the source of the number cannot be located based on the explanation;
    \item \textit{Correct reasoning, incorrect number}, the extracted number is incorrect, but the reasoning is sensible and supports the outcome. 
    \item \textit{Correct number, incorrect reasoning}, the extracted number is correct, but the reasoning process is incorrect, misleading, or does not support the outcome;
    \item \textit{Copy without reasoning}, the numbers are copied from the input papers with no understanding or reasoning steps leading to the final answer;
    \item \textit{Hallucinated}, the reasoning process refers to data, methods, or results not present in the input papers;
    \item \textit{Missing reasoning}, no reasoning process is generated.
\end{enumerate}

\section{Hyperparameters and APIs}\label{app:hyper_and_apis}
We executed all the experiments either via API or
on our own cluster. We used the paid-for OpenAI API to access GPT-3.5-turbo and GPT-4. On the
other hand, we hosted and trained the open-source models used in this paper on a distributed cluster.

SFT is performed for 5 epochs with a batch size of 1 (due to the 
large 
size of the input data) using a learning rate of $5 \times 10^{-5}$ and the AdamW optimiser~\cite{adamw}. For the RL setup, we adopt the GRPO algorithm~\cite{grpo}, training for 3 epochs with a learning rate of $1 \times 10^{-6}$, batch size 1, and 16 sampled generations per batch. Both training protocols leverage gradient accumulation with 8 accumulation steps. All experiments are conducted using the Open-R1 framework~\cite{openr1} on 8 NVIDIA A100 GPUs, each equipped with 80GB of memory. Models have been served for inference with the vllm framework~\cite{vllm}.

The weights of the final reward model have been set to maximise the importance of the correctness reward (0.8) while still ensuring the model follows the YAML format reward (0.1) and the already learnt thought format reward present in the base model (0.1). A tuning of such weights revealed that assigning equal weights to the 3 components leads to a degradation of the performance.  

\begin{figure*}
\centering
\includegraphics[width=0.98\textwidth]{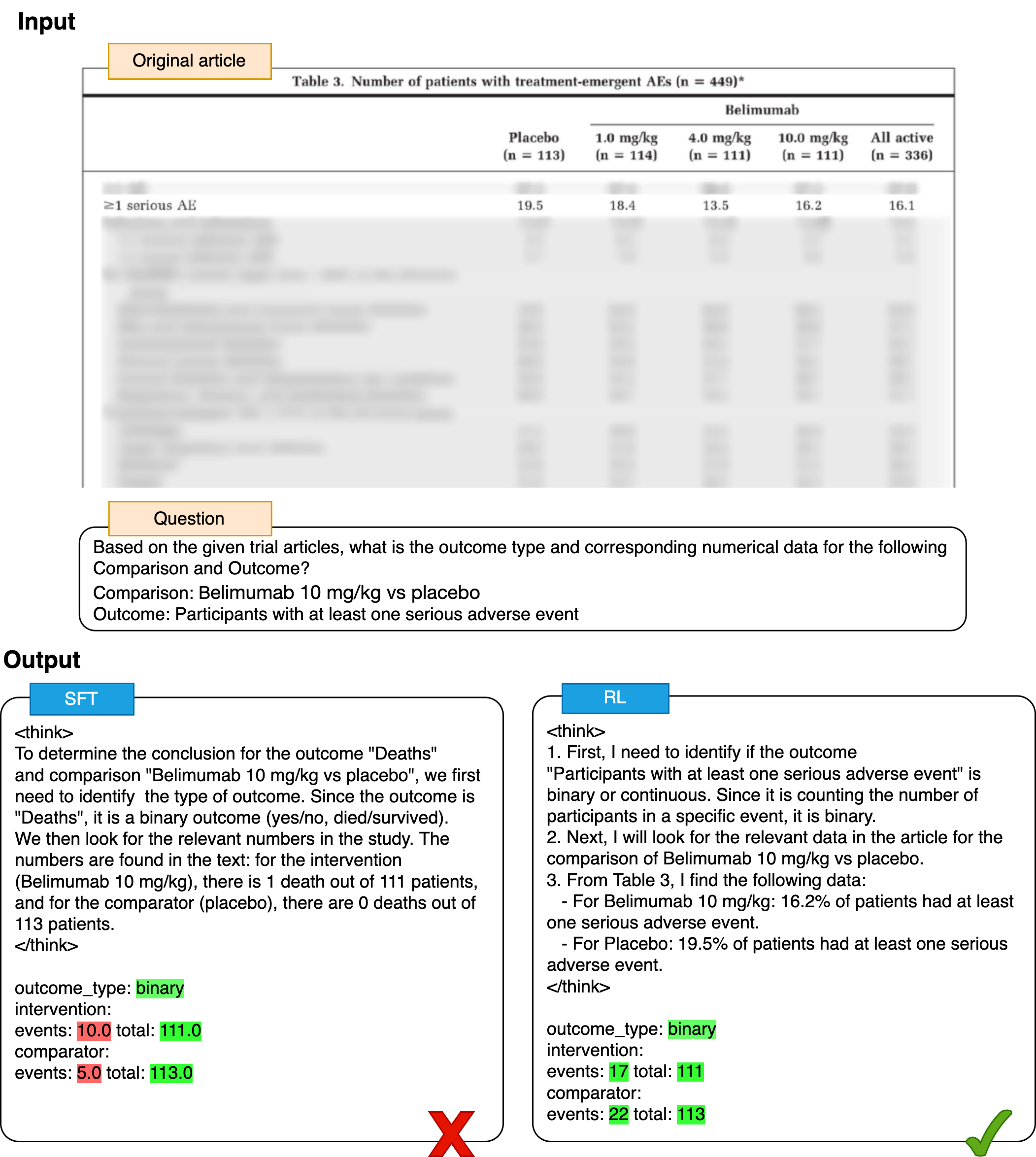}
\caption{Example of a complex example from \textsc{CochraneForest}. The SFT model fails at inferring the correct numerical information, and its thought mismatches its structured output. On the other hand, the model trained with GRPO correctly infers the number of events, reasoning on the previously extracted total number of participants. \textbf{Interestingly, even our human annotators initially struggled to extract the right numbers}.}
\label{fig:qualitative_example}
\end{figure*}

\section{Scientific Artefacts and Licensing}
In this work, we used the following scientific artefacts. LLaMa 3.1 is licensed under a commercial license\footnote{\url{https://llama.meta.com/doc/overview}}. GPT-4 is licensed under a commercial license\footnote{\url{https://openai.com/policies/terms-of-use}}. Qwen2.5 is licensed under the Apache 2.0 license\footnote{\url{https://qwenlm.github.io/blog/qwen3}}. DeepSeek models are licensed under the MIT license\footnote{\url{https://api-docs.deepseek.com/news/news250120}}. Mining text and data from the Cochrane library is permitted for non-commercial research through the Wiley API.\footnote{\url{https://www.cochranelibrary.com/help/access}}. The usage of the listed artefacts is consistent with their licenses.

 \begin{table*}[h!]
	     \centering
	     \begin{tabular}{l|>{\raggedright\arraybackslash}p{0.7\textwidth}}
		     \toprule
		     \rowcolor[gray]{0.9} \multicolumn{2}{c}{\textbf{Hawkey 2015}} \\
		     \midrule
		         \textbf{Text} & Importance Case reports and series suggest hematopoietic stem cell transplantation (HSCT) may benefit some patients with Crohn disease.Objective To evaluate the effect of autologous HSCT on refractory Crohn disease. ... \\ 
		         \midrule
		         \textbf{Outcome} & Clinical remission \\
                  \textbf{Comparison} & Stem cells versus Control\\
		         \textbf{Outcome type} & Binary \\
		         \textbf{Intervention events} & 8\\
		          \textbf{Intervertion total} & 23 \\
		           \textbf{Comparator events} & 2 \\
		          \textbf{Comparator total} & 22 \\
		         \textbf{Point estimate} & 3.83 \\
		         \textbf{95\% CI} & [0.91, 16.07]\\ 
		         \midrule
                 \textbf{Target value} & outcome\_type: binary\newline
		         intervention:\newline
		         events: 8 total: 23\newline
		         comparator:\newline
		         events: 2 total: 22\\
                 \midrule
                 \textbf{Thought} & To extract the relevant numbers, we first infer the outcome type as "binary" because the outcome "Clinical remission" is a yes/no or present/absent type of outcome. 
		         
		         The relevant numbers for this outcome are found in Table 2: "Final Assessments", under the outcome "CDAI <150 last 3 mo". 
                 Specifically, for the intervention group (Stem cells), we have 8 events (patients in clinical remission) out of 23 total patients (34.8\%). For the comparator group (Control), we have 2 events out of 22 total patients (9.1\%). \\
		     \bottomrule
		     \end{tabular}
	     \caption{The corresponding entry in \textsc{CochraneForestExt} for Hawekey 2015 in Figure~\ref{fig:fp_bin}. Thought and target value are contatenated for SFT as follows: \texttt{<think>\{thought\}</think>\{target\_value\}}.}
	     \label{tab:data_example}
	 \end{table*}
\end{document}